\def\year{2022}\relax
\documentclass[letterpaper]{article} 
\usepackage{aaai21}  
\usepackage{times}  
\usepackage{helvet} 
\usepackage{courier}  
\usepackage[hyphens]{url}  

\usepackage{natbib}  
\usepackage{caption} 
\usepackage{hyperref}       
\usepackage{stfloats}
\usepackage{url}

\usepackage{booktabs}       
\usepackage{nicefrac}       
\usepackage{microtype}      
\usepackage{amsthm}
\usepackage{amsfonts}
\usepackage{amsmath}
\usepackage{amssymb}
\usepackage{arydshln}
\usepackage{multirow}
\usepackage{graphicx, subfigure, float}
\usepackage{xcolor}

\definecolor{myred}{rgb}{0.67,0.1,0.1}

\title{Self-Initiated Open World Learning for Autonomous AI Agents}

\author {
	Bing Liu\textsuperscript{\rm 1},~Eric Robertson\textsuperscript{\rm 2},~Scott Grigsby\textsuperscript{\rm 2},~Sahisnu Mazumder\textsuperscript{\rm 1}  \\
}
\affiliations {
	\textsuperscript{\rm 1} Department of Computer Science, University of Illinois at Chicago, USA \\
	\textsuperscript{\rm 2} PAR Government Systems Corporation, USA\\
	 liub@uic.edu, \{eric\_robertson, scott\_grigsby\}@partech.com, sahisnumazumder@gmail.com
}

\begin{document}

\maketitle

{\large \color{myred}  An \textbf{extended and revised version} of this work has been published in the AI Magazine (AAAI) as follows: 
	
	\vspace{2mm}	
	\begin{quote}
		Bing Liu, Sahisnu Mazumder, Eric Robertson, and Scott Grigsby. ``\textbf{AI Autonomy: Self‐initiated Open‐world Continual Learning and Adaptation}." AI Magazine (2023). \\
		
		Click here for the pdf of the revised version: {\color{blue} \href{https://onlinelibrary.wiley.com/doi/pdf/10.1002/aaai.12087}{AI Magazine version link}}\\
	\end{quote}		
	
	\noindent
	\underline{\textbf{Please consider the AI Magazine 2023 version as }} \underline{\textbf{mentioned above for citation.}} \\
}

\begin{abstract}
As more and more AI agents are used in practice, it is time to think about how to make these agents fully autonomous so that they can learn by themselves in a self-motivated and self-supervised manner rather than being retrained periodically on the initiation of human engineers using expanded training data. As the real-world is an open environment with unknowns or novelties, \textit{detecting} novelties or unknowns, \textit{characterizing} them, \textit{accommodating or adapting} to them, \textit{gathering} ground-truth training data, and \textit{incrementally learning} the unknowns/novelties are critical to making the agent more and more knowledgeable and powerful over time. The key challenge is how to automate the process so that it is carried out on the agent's \textit{own initiative} and \textit{through its own interactions} with humans and the environment. 
Since an AI agent usually has a performance task, \textit{characterizing} each novelty becomes critical and necessary so that the agent can formulate an appropriate response to \textit{adapt} its behavior to \textit{accommodate} the novelty and to \textit{learn} from it to improve the agent's adaptation capability and task performance. The process goes continually without termination. This paper proposes a theoretic framework for this learning paradigm to promote the research of building \textit{Self-initiated Open world Learning} (SOL) agents. An example SOL agent is also described. 
\end{abstract}

\section{Introduction}
\label{introduction}
Classic machine learning (ML) makes~\textit{closed world assumption}, which means that what are seen by the system in testing or deployment must have been seen in training~\cite{fei2016breaking,bendale2015towards,liu2020learning}, i.e., there is nothing new or novel occurring in testing or deployment. This assumption is invalid in practice as the real world is an open environment with unknowns or novel objects. For humans, novelties or unknowns serve as an intrinsic motivation for learning. Human novelty detection results in a cascade of unique neural responses and behavioral changes that 
enable exploration and flexible memory encoding of the novel information. As learning occurs, this novelty response is soon lost as repeated exposure to novelty results in fast neural adaptation 
\cite{tulving1995novelty,murty2013hippocampal}. In order to make an AI agent thrive in the real open world, like humans, it has to detect novelties and learn them incrementally to make the system more knowledgeable and adaptable. It must do so \textit{on its own initiative} rather than relying on human engineers to retrain the system periodically. That is, it must learn in the \textit{open world} in a \textit{self-motivated manner} in the context of its \textit{performance task}.  


We use the hotel guest-greeting bot example from~\cite{ChenAndLiubook2018} to illustrate \textit{\textbf{S}elf-initiated \textbf{O}pen world \textbf{L}earning} (SOL). The \textit{performance task} of the bot is to greet hotel guests. When it sees a guest it has learned, e.g., John, it greets him by saying 

\vspace{+1mm}
``\textit{Hi John, how are you today}?''
\vspace{+1mm}

\noindent
When it sees a new guest, it should detect this guest as new or novel because it has never seen him/her before. This is a \textbf{\textit{novelty detection}} problem (also known as \textit{out-of-distribution} (OOD) \textit{detection}). It then needs to \textbf{\textit{accommodate}} or \textbf{\textit{adapt}} to the novel situation, e.g., by saying to the new guest

\vspace{+1mm}
``\textit{Hello, welcome to our hotel! What is your name, sir}?''
\vspace{+1mm}

\noindent
If the guest replies ``\textit{David},'' the bot takes some pictures of the guest to \textbf{\textit{gather training data}} and then \textit{\textbf{incrementally or continually learn}} to recognize David. The name ``\textit{David}'' serves as the \textit{class label} of the pictures taken. When it sees this guest again, it can say

\vspace{+1mm}
``\textit{Hi David, how are you today}?''~~~~ (he is no longer novel)
\vspace{+1mm}

Clearly, in an actual hotel, the situation is much more complex than this. 
First of all, if the bot uses a video camera, when it sees a novel object, it experiences a \textit{data distribution change} as it sees the new object for a period of time rather than just one image instance. It is desirable to detect the new object as quickly as possible. Next, how does the system know that the novel object is actually a person, not a dog? If the system can recognize the object as a person, how does it know that he/she is a hotel guest, not a policeman? In order to adapt to the novel object or situation, the system must first \textbf{\textit{characterize}} the novel object as without it, the agent does not know how to \textbf{\textit{adapt}}. In this case, some classification is needed to decide whether it is a person with luggage. If the object is a person but has no luggage, the bot will not respond or learn to recognize the person as it is \textit{irrelevant} to its performance task. If it is a moving object but not a person, it should notify a hotel employee and learn to recognize the object so that it will no longer be novel when it is seen again in the future. In short, for each characterization, there is a corresponding \textit{response} or \textit{adaptation strategy}, which can be NIL (i.e., do nothing). This discussion shows that in order to characterize, the agent must already have rich world knowledge. Clearly, there is also a \textbf{\textit{risk}} involved when making an incorrect decision. 



As classic learning matures, we should go beyond the existing framework to study how to enable the learner to learn by itself via its own interactions with humans and the environment, involving no human engineers.  
This paper proposes the SOL framework to promote the research of autonomous learning agents so that they can face the real open world and learn by themselves. An example SOL agent in the context of dialogue systems or chatbots that implements the main ideas in this paper will also be discussed. 

{\color{black}Although \textit{open world learning} has been studied by several researchers~\cite{bendale2015towards, fei2016learning,xu2019open}, it mainly focuses on novelty detection~\cite{parmar2021open}, which is also called \textit{open set detection} or \textit{out-of-distribution} \textit{detection} in the literature. Recently, zero-shot out-of-distribution detection is investigated in~\cite{esmaeilpour2021zero} based on the pre-trained model CLIP~\cite{radford2021learning}. Some researchers have further studied how to automatically identify the classes or categories of the detected novel object instances~\cite{shu2018unseen}. Yet, some others have studied learning the novel objects or classes after they have been detected~\cite{bendale2015towards,fei2016learning,xu2019open}. An recent survey of the topic can be found in~\cite{yang2021generalized}.
There is also a general discussion paper about open world learning in~\cite{langley2020open}, which presents many interesting blue sky ideas. However, SOL differs from these prior studies in many aspects. 

\textbf{(1)} SOL stresses self-initiation in learning, which means that all the learning activities from start to end are self-motivated and self-initiated by the agent itself. The process involves no human engineers. 

\textbf{(2)} Due to self-initiation, SOL enables learning after the model deployment like human learning on the job or while working, which has not been been attempted before. In existing learning paradigms, after a model has been deployed, there is no more learning until the model is updated or retrained on the initiation of human engineers. 

\textbf{(3)} SOL is also a lifelong and continual learning paradigm again because learning is self-initiated and can be done continually or lifelong. It is thus connected with lifelong and continual learning, which is another active research area in machine learning~\cite{ChenAndLiubook2018}. 

\textbf{(4)} SOL involves extensive interactions of the learning agent and human users, other AI agents, and the environment. The main purpose is to acquire ground-truth labels and training data by itself. 

\textbf{(5)} SOL makes learning autonomous and a learning agent equipped with SOL can face the real open world. We also believe that SOL is necessary for the next generation machine learning techniques. 
Finally, note that although SOL focuses on self-initiated learning, it does not mean that the learning system cannot learn a task given by humans or other AI agents incrementally or continually 
}



\section{Self-Initiated Open World Learning (SOL)} \label{sec:OWL}
Our agent has a \textit{primary performance task} and a pair of key modules $(T, S)$, where $T$ is the \textit{primary task-performer} (e.g., the dialogue system of the greeting bot) and $S$ is \textit{a set of supporting or peripheral functions} (e.g., the vision system and the speech system of the bot) that supports the primary task-performer. The primary task-performer as well as each support function may consist of four main sub-components ($L$, $E$, $N$, $I$), where $L$ is a SOL learner, $E$ is the executor that performs the task or function, $N$ is a novelty characterizer for characterizing each novelty so that $E$ can formulate an appropriate response to the novelty, and $I$ is an interactive module for the agent or $L$ to communicate with humans or other agents (e.g., to gain ground-truth training data). Since each function and the primary task performer has the same components, we will discuss them in general rather than distinguishing them, but will distinguish them when necessary. We will not discuss the relationships and interactions of the components. Below, we first discuss the SOL learner $L$, which starts with novelty and/or data shift detection. 


\subsection{Data Shift}
\label{sec.shift}
The classic ML depends on the \textit{independent and identically distributed} (IID) assumption. 
SOL deals with non-IID data, which is commonly know as \textit{data shift}. We use supervised learning as the task to develop the ideas, which can be easily adapted to other learning settings. 

Let the training data be $\mathcal{D}_{tr}=\{(\boldsymbol{x}_i, y_i)\}_{i=1}^{n}$, where $\boldsymbol{x}_i \in X$ is a training example following the training distribution $P_{tr}(\boldsymbol{x})$ and $y_i \in Y_{tr}$ is the corresponding class label of $\boldsymbol{x}_i$ and $Y_{tr}$ is the set of all class labels that appear in $D_{tr}$.

Note that $P(\boldsymbol{x},y)=P(y|\boldsymbol{x})P(\boldsymbol{x})$. Given $\boldsymbol{x} \in X$ and $y \in Y_{tr}$ in both training and testing, existing research has proposed the following three main types of data shift happening in testing~\cite{moreno2012unifying}. 

\textbf{Definition (covariate shift).} \textit{Covariate shift} refers to the distribution change of the input variable $\boldsymbol{x}$ between training and test phases, i.e., $P_{tr}(y|\boldsymbol{x})=P_{te}(y|\boldsymbol{x})$ and $P_{tr}(\boldsymbol{x}) \ne P_{te}(\boldsymbol{x})$, where $P_{te}$ is the test distribution.

\textbf{Definition (prior probability shift).}
\textit{Prior probability shift} refers to the distribution change of the class variable $y$ between training and test phases, i.e.,  $P_{tr}(\boldsymbol{x}|y)=P_{te}(\boldsymbol{x}|y)$ and $P_{tr}(y) \ne P_{te}(y)$.

\textbf{Definition (concept drift).} \textit{Concept drift} refers to the change in the posterior probability distribution between training
and test phases, i.e., $P_{tr}(y|\boldsymbol{x}) \ne P_{te}(y|\boldsymbol{x})$ and $P_{tr}(\boldsymbol{x})=P_{te}(\boldsymbol{x})$. 

Clearly, the most general data shift is: $P_{tr}(y|\boldsymbol{x}) \ne P_{te}(y|\boldsymbol{x})$ and $P_{tr}(\boldsymbol{x}) \ne P_{te}(\boldsymbol{x})$. 

However, there is another change not explicitly included in the above data shift, i.e., \textit{novelty} or \textit{novel instances} that appear in testing or deployment and do not belong to any known classes, which are the focus of SOL. Covariate shift may have novel instances belonging to known classes. 

\subsection{Novel Instances and Classes}

Novelty is an agent-specific concept. An object may be novel to one agent based on its partial knowledge of the world but not novel to another agent. In the context of supervised learning, the world knowledge is the training data $\mathcal{D}_{tr}=\{(\boldsymbol{x}_i, y_i)\}_{i=1}^{n}$ with $\boldsymbol{x}_i \in X$ and $y_i \in Y_{tr}$. Let $h(\boldsymbol{x})$ be the latent or internal representation of $\boldsymbol{x}$ in the agent's mind, $h(D_{tr}^i)$ be the latent representation of the training data of class $y_i$, and $k$ ($=|Y_{tr}|$) be the total number of training classes. We use $\mu(h(\boldsymbol{x}), h(D_{tr}^i))$ to denote the novelty score of a test instance $\boldsymbol{x}$ with respect to $h(D_{tr}^i)$. 
The degree of novelty of $\boldsymbol{x}$ with respect to $D_{tr}$, $\mu(h\boldsymbol{(x)}, h(D_{tr}))$, is defined as the minimum novelty score with regard to every class, 
\begin{equation}
\small
    \mu(h(\boldsymbol{x}), h(D_{tr})) = \min(\mu(h(\boldsymbol{x}), h(D_{tr}^1)), ..., \mu(h(\boldsymbol{x}),h(D_{tr}^k)))
    \label{TCLeq}
\end{equation}

The novelty function $\mu$ can be defined based on specific applications. For example, if the training data of each class follows the Gaussian distribution, one may use the distance from the mean to compute the novelty score. 

\textbf{Definition (novel instance)}: A test instance $\boldsymbol{x}$ is novel if its novelty score $\mu(\boldsymbol{x},.)$ is greater than or equal to a threshold value $\gamma$ such that $\boldsymbol{x}$ can be assigned a new class not in $Y_{tr}$. 

In \textit{covariate shift}, a novel instance may still be assigned to an existing class as the class assignment is application or agent specific. For example, we have a training class called \textit{animal} and the learner has seen \textit{dog} and \textit{chicken} in the \textit{animal} class, but during testing, a \textit{tiger} shows up. In the covariate shift case, \textit{tiger} can be added to the existing \textit{animal} class, although \textit{tiger} is novel as it has not been seen before. 






\textbf{Definition (novel class)}: A newly created class $y_{new}$ ($y_{new} \notin Y_{tr}$) assigned to some novel instances is called a \textit{novel class} (\textit{unknown} or \textit{unseen class}). The classes in $Y_{tr}$ are called \textit{known} or \textit{seen classes}.

Novelty is not restricted to the perceivable physical world but also includes the agent's internal world, e.g., novel interpretations of world states or internal cognitive states that have no correspondence to any physical world state. Interested readers may also read~\cite{boult2021towards} for a more nuanced and perception-based study of novelty. 


There are several related concepts to novelty. We clarify their difference from novelty here. 

\textbf{Outlier or anomaly}: An outlier is a data point that is far away from the main data clusters, but it may not be unknown. For example, the salary of a company CEO is an outlier with regard to the salary distribution of the company employees, but that is known and thus not novel. Unknown outliers are novel. Anomalies can be considered outliers or instances that are one off and never repeated. Though technically ``novel'' they may not need to result in a new class.

\textbf{Surprise or unexpectedness}: Based on the prior knowledge of the agent, the probability $P(\boldsymbol{x}|Q)$ of $\boldsymbol{x}$ occurring in a particular context $Q$  is very low, but $\boldsymbol{x}$ has occurred in $Q$, which is \textit{surprising} or \textit{unexpected}. If $\boldsymbol{x}$ has been seen before, it is not novel. In human cognition, surprise is an emotional response to an instance which greatly exceeds the expected uncertainty within the context of a task. Outliers, anomalies, and novelty can all lead to surprise.  



\subsection{Definitions of Learning in SOL}

The classic ML makes the \textit{i.i.d} assumption, which is often violated in practice. Here we define several other assumptions and learning paradigms that are progressively more and more aligned with the real-world learning needs of SOL.

\textbf{Definition (closed-world assumption)}: 
No new or novel classes appear in testing. Other types of data shift may occur.  

\textbf{Definition (closed-world learning)}: It refers to the learning paradigm that makes the closed-world assumption. 

\textbf{Definition (learning with data shift)}: It refers to the learning paradigm that deals with certain types of data shift in testing or deployment, but not new classes. 



\textbf{Definition (open world learning)}: It refers to the learning paradigm that can detect data shifts and novel instances in testing or deployment. The learner can learn the novel classes labeled by humans from the identified novel instances and update the model using the new data. The re-training or model updating is initiated by human engineers. That is, there is no self-initiated learning after model deployment. 

Note that for covariate shift, the assignment of shifted data (or novel) instances to existing training classes is normally done by humans. For concept shift, the shifted classes are caused by humans or by the environment. 

\textbf{Definition (self-initiated open world learning (SOL)):} {\color{black}It refers to the learning paradigm that has the capability of open-world learning and the agent is able to initiate a learning process by itself during application (after deployment) with no involvement of human engineers. The learning of the new classes is incremental, i.e., no re-training of previous classes. The process is thus lifelong or continuous, which makes the agent smarter and smarter over time.} 

Note that apart from learning, SOL's performance task also demands characterization and adaptation, which we discuss in Section~\ref{sec.char}. 

\subsection{Steps in Learning in SOL}
\label{sec.steps}

Learning in SOL involves the following three main steps. We ignore updating the existing model to deal with data shift because if we can deal with novel classes, it is relatively easy to deal with the other types of data shift (see Section~\ref{sec.shift}). 

\vspace{+1mm}
\noindent
\textbf{Step 1} - \textit{Novelty detection}. This step involves detecting data instances (1) whose classes do not belong to $Y_{tr}$ or (2) have covariate shift, and it is done automatically. A fair amount of research has been done on this under open-set classification or out-of-distribution detection~\cite{pang2021deep}.  

\vspace{+1mm}
\noindent
\textbf{Step 2} - \textit{Acquiring class labels and creating a new learning task on the fly}: This step first clusters the detected novel instances. Each cluster represents a new class. It may be done automatically or through interactions with humans using the interaction module $I$. Interacting with humans should produce more accurate clusters and also obtain meaningful class labels. If the detected data is insufficient for building an accurate model to recognize the new classes, additional ground-truth data may be collected via interaction with humans. A new learning task is then created. 

In the case of our hotel greeting bot, since the bot detects a single new guest (automatically), no clustering is needed. It then asks the guest for his/her name as the class label. It also takes more pictures as the training data. With the labeled ground-truth data, a new learning task is created to incrementally learn to recognize the new guest on the fly.  

The learning agent may also interact with the environment to obtain training data. In this case, the agent must have an \textbf{\textit{internal evaluation system}} that can assign rewards to different states of the world, e.g., for reinforcement learning. 

\vspace{+1mm}
\noindent
\textbf{Step 3} - \textit{Incrementally learn the new task.} After ground-truth training data has been obtained, the learner $L$ incrementally learns the new task. This is \textit{continual learning}~\cite{ChenAndLiubook2018}, an active and challenging ML research area, which is defined as learning a sequence of tasks incrementally. 





\subsection{Novelty Characterization and Adaptation} \label{sec.char}
In a real-life application, classification may not be the primary task. 
For example, in a self-driving car, object classification supports its primary performance task of driving. 
To drive safely, the car has to take some actions to adapt or respond to the new object, e.g., slowing down and avoiding the object. In order to know what actions to take, it must characterize the new object. {\color{black}Based on the characterization, appropriate actions are formulated to accommodate or respond to the novel object. The response process may also involve learning.} 











\textbf{Definition (novelty and response)}: It is a pair $(c, r)$, where $c$ is the \textit{characterization of the novelty} and $r$ is the \textit{response} to the novelty, which is a plan of dynamically formulated actions based on the characterization of the novelty and the agent’s interactions with the novel item. If the system cannot characterize a novelty, it will take the \textit{default response}. 
In some situations, the agent does not know what to do. The response is \textit{Learn}, i.e., to learn the actions to take.

\textbf{Definition (characterization of novelty)}: {\color{black}It is a description of the novelty based on the agent's existing knowledge of the world. According to the description, the agent chooses a specific course of actions to respond to the novelty. }

Characterization of novelty can be done at different levels of detail, which may result in more or less precise responses. Based on the ontology and attributes related to the performance task, the description can be defined based on the \textit{type of the object} and the \textit{attribute of the object}. For example, for self-driving cars, when sensing a novel object, the car can identify both the movement of the object and the location of the object relative to the direction of travel--on the road or off the road. Thus, some classification of movement and location in this case is needed to characterize the novelty which, in turn, facilitates determination of the agent's responding action(s). For instance, if the novel object is a mobile object, the car may wait for the object to leave the road before driving. 

Another common characterization strategy is to \textit{compare the similarity }between the novel object and the existing known objects. For example, if it is believed that the novel object looks like a pig, then the agent may react like when it sees a pig on the road. 

In our greeting bot example, when it can characterize a novelty as a new guest, its response is to say "\textit{Hello, welcome to our hotel! What is your name, sir}?" 
If the bot has difficulty with object characterization, it can take a \textit{default action}, either `{do nothing}' or `{report to a hotel staff}.' The set of responses are specific to the application. For a self-driving car, the default response to a novel object is to slow down the car as soon as possible so that it will not hit the novel object.

This discussion implies that in order to effectively characterize a novelty, the agent must already have a great of world knowledge that it can use to describe the novelty. Without such knowledge, it will not be able to characterize the novelty in meaningful ways. 

The characterization and response process is often interactive in the sense that based on the initial characterization, the agent may choose a course of actions, but after some actions are taken, it will get some feedback from the environment. Based on the feedback and the agent's additional observations, the course of actions may be revised. 

{\color{black}\textbf{Focus of attention.} Due to the performance task, the agent should focus on detecting and characterizing novelties that are critical to the performance task. For example, in the self-driving car application, the agent should focus on novel objects or events that are or may potentially appear on the road in front of the car. It should not pay attention to novel objects in the shops along the street as they do not affect driving. In characterizing a novel object on the road, it should focus on those aspects that are important to driving, i.e., whether it is a still or a moving object. If it is a moving object, the agent must determine its direction of moving.}  

\textbf{Learning to respond.} As indicated above, in some situations, the system may not know how to respond to a new object or situation. In this case, it needs to learn by itself (see Section~\ref{sec.steps}). There are many ways to learn, e.g., 

(1) \textit{Asking a human}. In the case of the self-driving car, when it does not know what to do, it may ask the passenger using the interactive module $I$ (e.g., in natural language) and then follow the instruction and also remember or learn it for future use. For example, if the car sees a black patch on the road that it has never seen before, it can ask ``\textit{what is that black thing in front?}'' The passenger may answer ``\textit{that is tar}.''  In the case of an unrecognized response, such as no prior information on tar, the system may progress with further inquiry, asking the passenger ``\textit{what should I do?}''  

(2) \textit{Imitation learning}. On seeing a novel object, if the car in front drives through it with no issue, the car may choose the same course of action as well and learn it for future use. 

(3) \textit{Performing reinforcement learning}. By interacting with the environment through trial and error exploration, the agent learns a good response policy for future use. As mentioned earlier, in this case the agent must have an internal evaluation system that can assign rewards to environment states. 

If multiple novelties are detected at the same time, it is more difficult to respond as the agent must reason over the characteristics of all novel objects to dynamically formulate an overall plan of actions that prioritize the responses. 

\subsection{Risk}
There is risk in achieving performance goals of an agent when making an incorrect decision. For example, classifying a known guest as unknown or an unknown guest as known may negatively affect guest impressions resulting in negative reviews. For a self-driving car, misidentifications can result in wrong responses, which could be a matter of life and death. Thus, risk assessment must be made in making each decision. Risk assessment can also be learned from experiences or mistakes.  In the example of a car passing over tar, after the experience of passing over shiny black surfaces safely many times, if the car slips in one instance, the car agent must assess the risk of continuing the prior procedure.  Given the danger, a car may weight the risk excessively, slowing down on new encounters of shiny black surfaces.

\vspace{+1mm}


\section{An Example SOL System}
Although novelty detection~\cite{yang2021generalized,pang2021deep} and incremental or continual learning~\cite{ChenAndLiubook2018,Parisi2019continual} have been studied widely, little work has been done to build a SOL system. 
Here we describe a dialogue system (called CML) that is based on the SOL framework and performs each function in SOL continually by itself after the system has been deployed~\citep{mazumder2020application,liu2021lifelong}. 

CML is an natural language interface like Amazon Alexa and Siri. Its \textbf{\textit{performance task}} is to take a user command in natural language (NL) and perform the user requested API action in the underlying application. There is no support function in this application. The key issue is how to understand paraphrased NL commands from the user in order to map a user command to a system’s API call. \textit{\textbf{Novelty} }equates to the system's failure in understanding a user command. After the system automatically detects a novelty (a hard-to-understand user command), it will try to understand the command and also learn the command so that it will be able to understand it and related commands in the future. The novelty \textbf{\textit{characterization}} step of SOL here is to identify the part of the command that the system does not understand. Based on the characterization, the system \textbf{\textit{adapts}} by asking the user via an interactive dialogue to obtain the ground truth API action requested by the user, which also serves as a piece of training data for \textbf{\textit{continual learning}}. In the adaptation or accommodation process, \textit{\textbf{risk}} is also considered.   

Consider the following example. The user issues the command ``\textit{turn off the light in the kitchen}'' that the system does not understand (i.e., a \textbf{novelty}), Based on the current system state, it decides which part of the command it can understand and which part it has difficulty (i.e., \textbf{characterization}). Based on the characterization result, it provides the user a list of \textit{top-k} predicted actions (see below) described in NL and asks the user to select the most appropriate action from the given list (i.e., \textbf{adaptation}). 
    
\begin{quote}
    \vspace{-2mm}
    \small
    \rule{0.9\linewidth}{0.4pt}\\
    \textbf{Bot}: Sorry, I didn't get you. Do you mean to:\\
    \vspace{-2mm}\\
    \textbf{option-1.}~~ switch off the light in the kitchen, or\\ \textbf{option-2.}~~ switch on the light in the kitchen, \\ 
    \textbf{option-3.}~~ change the color of the light?\\
    \rule{0.9\linewidth}{0.4pt} 
    \vspace{-1mm}
\end{quote}
    
The user answers the desired action (option-1). The action API [say, \texttt{SwitchOffLight}(\textit{arg}:place)] corresponding to the selected action (option-1) is retained as the ground truth action for the issued NL command. In subsequent turns of the dialogue, the agent can also ask the user questions to acquire ground truth values associated with arguments of the selected action, as defined in the API. CML then \textbf{incrementally learns} to map the original command ``\textit{turn off the light in the kitchen}'' to the API action,  \texttt{SwitchOffLight}(\textit{arg}:place). This learning ensures that in the future the system will not have problem understanding the related commands.  

\textbf{Risk} is considered in the system in two ways. First, it does not ask the user too many questions in order not to annoy the user. Second, when the characterization is not confident, the system simply asks the user to say his/her command again rather than providing a list of random options to choose from.




\section{Key Challenges}
Although novelty detection and continual learning have been researched extensively~\cite{yang2021generalized,pang2021deep}, they remain to be challenging. Limited work has been done to address the following (this list is by no means exhaustive): 

\textbf{Obtaining training data on the fly.} One key feature of SOL is interaction with humans to obtain ground-truth training data, which needs a dialogue system. 
Building an effective dialogue system is very challenging. We are unaware of any such system for SOL except CML~\cite{mazumder2020application}, but CML is only for simple command learning. 

\textbf{Few-shot continual learning.} It is unlikely for the learning agent to collect a large volume of training data via interaction with the user. Then, an effective and accurate few-shot continual/incremental learning method is necessary. 

\textbf{Novelty characterization.} This is critical because it defines the characteristics used to recognize world state and determine the best response strategy. For example, if a self-driving car encounters a novel/new mobile object, its response will be different from encountering an immobile object. Even if we know it is a mobile object, the system may also need to know which direction it is moving and the moving speed in order to formulate an appropriate response. 
The challenge is that a large number of classifiers or other models may need to be built. This means that the system or agent must have a very rich world model and a large amount of knowledge related to its performance task before it can effectively characterize novelties. 


\textbf{Learning to respond.} This task is especially challenging in a real-world physical environment~\cite{dulac2021challenges}. 
For example, due to safety concerns, learning during driving (which is required by SOL) by a self-driving car using reinforcement learning (RL) is very dangerous. {\color{black}Furthermore, for RL to work, the agent must have a highly effective internal reward or evaluation system to assign rewards to actions and states and to be aware of \textit{safety constraints}. Little work has been done so far.  



\textbf{Knowledge revision.} It is inevitable that the system may misinterpret, generalize or otherwise assemble incorrect knowledge.  A SOL system must have a mechanism to detect and revise the inaccurate knowledge on its own. Again, little work has been done. }

\section{Conclusion}
\label{conclusion}

A truly intelligent system must be able to learn autonomously and continually in the open world on its own initiative after deployment in order to adapt to the ever-changing world and gain more and more knowledge to become more and more powerful over time. This paper proposed a \textit{self-initiated open world learning} (SOL) framework for the purpose, and presented the concepts, steps and key challenges. An example SOL system is also described. 
We believe that the future research in SOL will bring ML and AI to the next level. 

\section*{Acknowledgments}
This paper benefited greatly from numerous discussions in the DARPA SAIL-ON Program PI meetings. This work was supported in part by a DARPA Contract HR001120C0023. Bing Liu and Sahisnu Mazumder are also partially supported by two National Science Foundation (NSF) grants (IIS-1910424 and IIS-1838770), and a Northrop Grumman research gift. The views expressed
in this document are those of the authors and
are not those of funders. 


\bibliography{references}
\bibstyle{aaai21.bst}

\end{document}